\documentclass[twocolumn]{bmcart}

\usepackage{graphicx}
\usepackage{amsthm,amsmath,algorithm2e}
\usepackage{amssymb}

\RequirePackage{natbib}
\usepackage[utf8]{inputenc}		
\usepackage[document]{ragged2e}	

\startlocaldefs
\endlocaldefs

\begin{document}

\begin{frontmatter}

\begin{fmbox}
\dochead{Research}


\title{Static Force Field Representation of Environments Based on Agents' Nonlinear Motions}


\author[
addressref={aff1},                   
corref={aff1},                       
email={damian.campo@ginevra.dibe.unige.it}   
]{\inits{DC}\fnm{Damian} \snm{Campo}}
\author[
addressref={aff1, aff2},
email={alejandro.betancourt@ginevra.dibe.unige.it}
]{\inits{AB}\fnm{Alejandro} \snm{Betancourt}}
\author[
addressref={aff1},
email={lucio.marcenaro@unige.it}
]{\inits{LM}\fnm{Lucio} \snm{Marcenaro}}
\author[
addressref={aff1},
email={carlo.regazzoni@unige.it}
]{\inits{CR}\fnm{Carlo} \snm{Regazzoni}}

\address[id=aff1]{
	\orgname{University of Genova, Department of Naval, Electric, Electronic and Telecommunications Engineering}, 
	\street{Via all'Opera Pia 11},
	\city{Genova},                            
	\cny{ITA}                                 
}
\address[id=aff2]{
	\orgname{Eindhoven University of Technology, Designed Intelligence Group, Department of Industrial Design},
	\city{Eindhoven},
	\cny{NDL}
}





\begin{abstractbox}
	
	\begin{abstract} 
%
\justify
This paper presents a methodology that aims at the incremental representation of areas inside environments in terms of attractive forces. It is proposed a parametric representation of velocity fields ruling the dynamics of moving agents. It is assumed that attractive spots in the environment are responsible for modifying the motion of agents. A switching model is used to describe near and far velocity fields, which in turn are used to learn attractive characteristics of environments. The effect of such areas is considered radial over all the scene. Based on the estimation of attractive areas, a map that describes their effects in terms of their localizations, ranges of action and intensities is derived in an online way. Information of static attractive areas is added dynamically into a set of filters that describes possible interactions between moving agents and an environment. The proposed approach is first evaluated on synthetic data, posteriorly, the method is applied on real trajectories coming from moving pedestrians in an indoor environment. 

	\end{abstract}
		
	
	\begin{keyword}
		\kwd{Kalman filtering}
		\kwd{Interactive force models}
		\kwd{Trajectory analysis}		
		\kwd{Representation of environments}
		\kwd{Situation awareness}
	\end{keyword}
	
	
\end{abstractbox}
\end{fmbox}

\end{frontmatter}




\section{Introduction}
\justify
Analysis of trajectories performed by moving entities in environments is an important topic for different fields such as video surveillance \cite{Hsieh2008}, crowd/vehicle analysis \cite{Mahadevan2010, Akoz2014} and in general for monitoring systems, on which the dynamics of agents can lead to a better understanding of patterns and situations of interest \cite{Morris2011, Wang2011}. Abnormality detection is one of the most explored applications that involves analysis of trajectories. In such approach, by characterizing agents' motions, it is possible to learn and identify normal/abnormal situations in a certain environment.

In general, approaches for abnormality detection are based on a set of observations that define the regular behaviors in a scene. Afterwards, abnormalities are defined as behaviors that do not match with patterns previously learned as normal, i.e., behaviors that have not been observed before \cite{Kim2011}. Accordingly, observations that deviate from the characterized normal patterns are classified as potential abnormalities. Abnormality detection has been successfully used in real applications such as identification of hazardous massive crowds \cite{Helbing2000,Li2015}, recognition of group activities \cite{Vahid2015,Zhou2015}, traffic event classification \cite{Hu2004, Akoz2014, Bastani2016, Haag2000, Lan2016}, detection of anomalies in maritime trajectories \cite{Castaldo2014,Lei2016}, among others.

The recognition of motion patterns and interactions from trajectories is a challenging task, in particular when dynamics of moving agents are nonlinear through time and space. Moreover, a quick analysis of existent bibliography shows that current methods usually claim robustness and reliability for highly restricted scenarios, broad sensor availability and short video footages \cite{Remagnino2007}. To alleviate these issues, a common approach is to represent the state of agents under a probabilistic framework and characterize the motion patterns by following a nonlinear Bayesian state estimation. This approach has been successfully applied in surveillance environments and complex motion patterns \cite{Bastani2016, Nascimento2010}.

In the present work, abnormalities from a proposed baseline model are used to parameterize characteristics of environments by looking at trajectories of individuals in an online way.    

The majority of approaches that deal with analysis of trajectories are focused on characterizing agents' dynamics through time \cite{Andersson2013, Lu2016, Cancela2013} or learning interactive forces exerted by multiple moving agents. The latter can be understood as an evolution of the seminal paper of Helbing and Molnár \cite{Helbing1995}, on which crowd motions are described by a combination of simple interactive forces between pedestrians. The methodology proposed by the authors is based on social force models, a concept originally introduced by \cite{lewin1951}. In a recent work, Seer et al. \cite{SEER2014} extend the work of Helbing by using three different shapes of social forces models, namely radial \cite{Helbing1995}, elliptical \cite{Helbing2009} and a split version of forces \cite{Rudloff2011}. The work of Seer is evaluated on real pedestrian trajectories.

A common trend in existent works, is the characterization of moving agents' trajectories based on their interactions among them, without taking into account the environment and its effects in the state of agents. Our work is motivated by the approach of Seer \cite{SEER2014}, with the main difference that the motion of our agents is characterized by environment forces revealed hierarchically due to interactions of the type agent-static zone.

This work tackles the analysis of moving agents in cases where the only information available is their location through time. The proposed strategy can be used in places on which a detailed map is unknown but agents' positions can be obtained via GPS or other localization systems like Wi-Fi fingerprint for indoor environments \cite{Liu2012,Montorsi2014}. Furthermore, the proposed approach provides information of the moving agents' surroundings that can be useful in cases of places monitored by cameras that provide noisy measurements, partial/full occlusion or scenes with illumination changes \cite{Yilmaz2006}.

The present work proposes a bank of Kalman filters (KFs) that explains the motion of agents in a hierarchical way. Models obtained from the bank of filters are incrementally learned by looking at abnormalities or deviations from a reference formulation based on a random walk behavior, such idea was first presented in \cite{Damian2016}, where an off-line non-parametric method was proposed to characterize effects of static external objects in simulated environments.

In the current work, dynamical models that describe agents' motions can be used to hypothesize a semantic representation of the environment effects \cite{Wang2011, Lou2002}. In other words, the agents' motions incrementally reveal characteristics of the environment. By considering a Bayesian approach, the proposed method learns sequentially the nonlinear dynamics of moving agents over segments in which their orientations are relatively stable. Consequently, such segments can be seen as letters that are part of a vocabulary learned dynamically. As a future work, these vocabularies are proposed to be used for identifying normal and abnormal situations related to environment's characteristics revealed on the fly.

The novelties of this work are itemized as follows: i) It is proposed a Bayesian approach for understanding the physical and non-physical surrounding areas of moving agents based on their nonlinear dynamics. The proposed formulation allows to characterize parts of environments individually by using a bank of filters that encodes information about learned models in an incremental way. ii) A parametric force field model based on a switching process is proposed to understand agents' motions. Such model includes information about how static areas influence the dynamics of moving agents. Additionally, only agents' positions are necessary to understand the effect produced by static areas through time. iii) The proposed methodology does not need a broad sensor availability, since only location information is needed, any type of trajectory can be analyzed and modeled. Our approach is evaluated on synthetic data and real trajectories of pedestrians in an indoor place.

The rest of this article is organized as follows: A description of the agents considered in this work is provided in Section \ref{AgentsDef}. Force model and problem definition are described in Section \ref{ProblemDef}. A hierarchical representation of the environment is explained in Section \ref{SemanticRep}. Results obtained with synthetic and real data are given in Section \ref{Results}; and Section \ref{Conclusions} concludes the article.

\section{Agent paradigm}\label{AgentsDef}
\begin{figure}[h!] 
	\includegraphics[width=8cm]{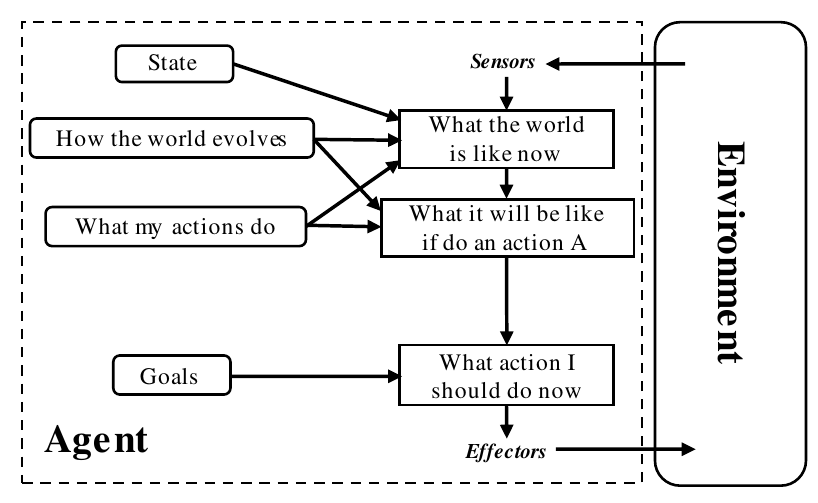}
	\caption{\csentence{Goal-based agent.}
		Schematic diagram of an agent with explicit goals proposed by \cite{Russell2003}.}
	\label{fig:fig1}
\end{figure}

\justify
This work proposes a method for identifying and characterizing attractive areas in environments based on observations of cognitive entities that move through them. Accordingly, the present approach assumes that such moving entities can be represented as goal-based agents described in the work of Russell and Norvig \cite{Russell2003}. Decisions made by such class of agents are based on a cognitive perception of their surroundings combined with a goal to be achieved. As can be seen in the diagram shown in Fig.\ref{fig:fig1}, the agent's goal plays a fundamental role at the moment of taking a decision.

In this work, agents' goals are considered to be areas in the environment where agents tend to go. For modeling such areas, it is considered to use a formulation based on social force models. An idea introduced by Helbing and Molnár in \cite{Helbing1995} for understanding pedestrian dynamics. Consistently, each main effect described in their work is here contextualized with the final purpose of explaining effects produced by unknown attractive areas in environments. 

Helbing and Molnár propose that there are three main effects with which the motion of single pedestrians can be explained: \textit{i) Destination to reach}: It consists in the shortest path to follow in order to go towards a goal. \textit{ii) Influences caused by other pedestrians and barriers}: In which they define a repulsive force model for describing the effect that pedestrians exert among others at the moment of avoiding people. They also propose a repulsive model to explain the effect of barriers, such as borders of buildings or obstacles. \textit{iii) Temporary attractive effects}: They define a model that explains the sudden attractiveness that could be experienced by a pedestrian, it is the case of feeling attraction to other people (friends, street artists, etc.) or going towards static objects such as window displays.

\begin{figure}[h!] 
	\includegraphics[width=8cm]{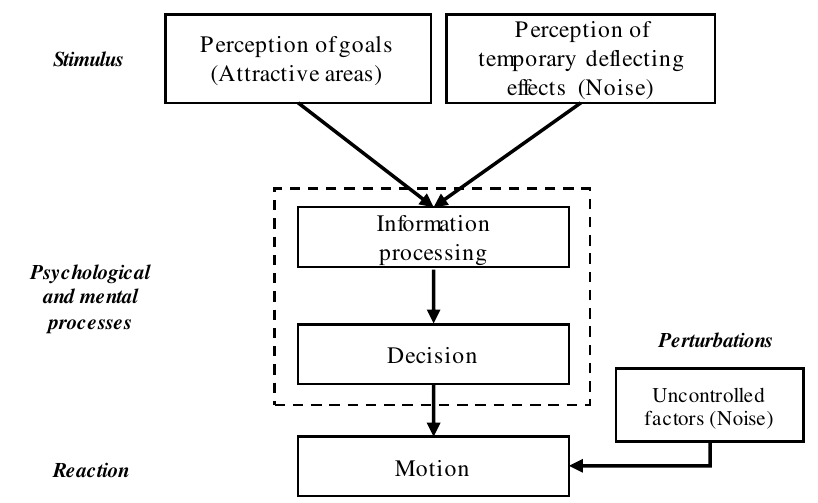}
	\caption{\csentence{Schematic representation of processes that induce behavioral changes.}
		Proposed scheme to represent the motions of agents based on \cite{Helbing1995}.}
	\label{fig:fig2}
\end{figure}

Based on the three main effects described previously for understanding pedestrian dynamics, it is proposed to use the same reasoning to understand goal-based agents dynamics. From this viewpoint, main goals are assumed to be represented as \textit{i) Destinations to reach} that can be modeled as attractive fields in the environment. \textit{ii) Influences caused by other pedestrians and barriers} and \textit{iii) Temporary attractive effects} are proposed to be modeled as noise that deviates agents from their main destinations. Accordingly, the diagram in Fig.\ref{fig:fig2} explains the proposed model to understand the dynamics of moving agents in a scene. 

In literature, the majority of works that deal with analysis of trajectories are focused on modeling and learning agents' motions through time \cite{Morris2011,Bastani2016,Andersson2013,Lei2016,Lu2016,Cancela2013}. However, in the proposed methodology, since an interaction between agents and the environment is assumed, it is useful to consider an attractive force model for explaining agents' dynamics. Such model includes information about environment properties that define how moving agents are influenced by static areas in the scene. Accordingly, when an agent's motion pattern is observed, the understanding of environment static characteristics is enriched.

By considering that dynamics of agents can be seen as the result of external forces acting on them, it is possible to associate such forces with the internal motivations that make agents perform certain actions (movements) \cite{Helbing1995} in order to reach a particular destination. From this perspective, each destination point is proposed to be modeled as an attractive area that exerts a force field in the environment.

The methodology proposed in this paper aims at the incremental understanding of attractive effects that static areas exert over moving agents. This work can be seen a method to improve the situation awareness (SA) of an environment by looking at moving entities. Formally, SA is defined as "the perception of elements within a volume of time and space, the comprehension of their meaning, the explanation of their present (observed) status and the ability to project the same in near future instants." \cite{Bhatt2012}. Subsequently, the methodology proposed in this paper increases the SA of environments incrementally as agents' motions are observed.

As an individual agent model is proposed, the current method aims at the understanding of agents' surroundings in terms of attractive areas (goals) in cases where interactions with other agents are momentary, i.e., in situations where interactions between individuals can be modeled as a noise parameter. In cases of multi-agent systems composed by a large density of individuals as proposed in \cite{Masoud2007}, the main goal of agents can be confused with the constant evasion of multiple individuals and the proposed approach would not work properly. Based on that, our method is focused on cases where there is a low density of agents that interact with each other.  

\section{Force model and problem definition}\label{ProblemDef}
Taking into consideration a classical mechanics approach, a force is defined as a vectorial quantity that acts on a body to cause a change in its state of motion \cite{Wagh2012}. Forces can be classified in action-reaction pairs: When bodies, which are in contact, change their momentum \cite{Wagh2012} and action-at-a-distance forces: When agents interact without being physically touched \cite{hesse2005}. Throughout this work, motivations of agents are modeled as action-at-a-distance forces that can be characterized by looking at the motion of individuals. 

A force field $\vec{F}$ is defined as a vector point-function which has the property that at every point of the space takes a particular value related to the magnitude and direction of a force acting on a particle placed there and whose mass is considered as one \cite{Tenenbaum2012}. Accordingly, moving agents' dynamics are used to learn the effects produced by force fields in the environment. 

A central force field $\vec{F} = f(r) \hat{r}$ is a particular case of fields where the motion of agents is affected depending on the distance $r$ to the location of an attractive/repulsive force source, called center of force. $\hat{r}$ is a unit vector that points in the direction of $r$. Based on that, this article proposes radial force fields for modeling dynamics of agents through time.

For characterizing the effect that an attractive area exerts over moving agents, it is considered the second Newton's law $\vec{F} = m\vec{a}$, where $m = 1$ as proposed before. Therefore, it is direct to infer that $\vec{F} = \vec{a}$, i.e., the force exerted by an attractive area in the environment can be measured as the agent's acceleration at a particular point where it is located.  

It is hypothesized that information about the agent's speed $V(t)$ can be inferred through time $t$. Then, it is possible to relate agents' speeds with their accelerations as can be seen in equation \eqref{eq1}: 
\begin{equation}\label{eq1}
V_{t_{2}} = \int_{t_{1}}^{t_{2}} a_{t_{1}} dt = a_{t_{1}} (t_{2} - t_{1}) + c,
\end{equation}
where $t_{1}$ and $t_{2}$ are two consecutive time instants, $a_{t_{1}}$ represents the force magnitude that acts over an agent at time $t_{1}$ and $c$ is an integration constant that represents the agent's velocity $V_{t_{1}}$ at the instant $t_{1}$. By considering a number of $n$ spatial dimensions from which trajectories of moving agents are observed, the location of a single agent can be expressed as shown in equation \eqref{eq1a}:
\begin{equation}\label{eq1a}
\mathcal{X} = (d_1, d_2,...,d_n).
\end{equation}

The velocity field experienced by an agent influenced by a central force field in terms of its location $\mathcal{X}$ is shown in equation \eqref{eq2}:
\begin{equation}\label{eq2}
\vec{V}_{\mathcal{X}(t)} = \vec{a}_{\mathcal{X}(t-1)}  \Delta t + \vec{V}_{\mathcal{X}(t-1)}, 
\end{equation}
where it is possible to observe that dependencies of time $t$ can be substituted by spatial coordinates $\mathcal{X}$. Since force fields are fixed in one location and their influences are assumed to remain equal through time, agents' dynamics will depend only on their own location $\mathcal{X}$. 

A switching model is proposed to characterize the velocity fields perceived by agents due to the presence of attractive radial force fields. Based on this, two motion stages are defined: i) When the agent is near to an attractive center of force (near range of interaction) and ii) When the agent is far from it (far range of interaction). 

\subsection{Near range of interaction}\label{NearModels}
In order to model the way by which moving agents arrive in attractive areas, it is considered a parametric function that describes their velocities depending on the distance $r$ to a center of force. The proposed model is based on a top speed from which the agent continuously experiments a deceleration according to a function inspired by the repulsive terms offered by \cite{Helbing1995, Helbing2009} and the deceleration factor described on \cite{Rudloff2011}. 

\begin{figure}[h!] 
		\includegraphics[width=8cm]{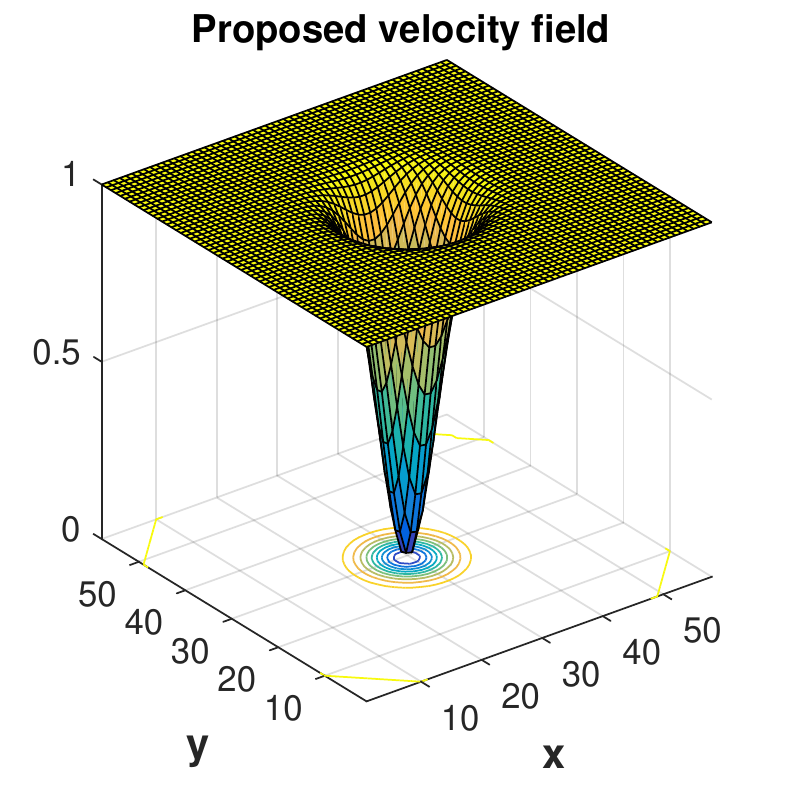}
	\caption{\csentence{Attractive near velocity field representation.}
		The proposed shape of the velocity field experienced by agents when they are approaching to an attractive area is illustrated.}
	\label{fig:fig3}
    \end{figure}

Consistently, for characterizing the arrival of agents in attractive areas, it is proposed a velocity field $\vec{G}_{near}(r)$ that defines how their dynamics vary as they approach closer to an attractive area. Two global terms can explain such velocity field, one related to a baseline velocity and other based on a deceleration term, as described in equation \eqref{eq3}:
\begin{equation}\label{eq3}
\vec{G}_{near}(r)= \bigg(\underbrace{\beta}_{\substack{\text{Baseline} \\ \text{velocity}}} - \underbrace{\alpha e^{\frac{-r^2}{\sigma^{2}}}}_{\substack{\text{Deceleration}\\\text{term}}}\bigg) \hat{r},
\end{equation}
where $r$ is defined as the Euclidean distance between an agent and the attractive area in question, such that $r = ||\mathcal{X}-\mathcal{X}_{0}||_2$, where $\mathcal{X}$ represents the agent's location and $\mathcal{X}_{0}$ is the unknown position of the attractive area. $\sigma$ is an unknown parameter that defines the way in which moving agents decelerate while they approach the center of force $\mathcal{X}_0$. Low values of $\sigma^{2}$ are associated with abrupt agent decelerations; and high values are related to smooth decrements of speed while agents advance towards $\mathcal{X}_{0}$. $\beta$ is the top speed from which agents start decelerating until they arrive at the attractive area. $\alpha$ defines the final speed that an agent will have when arrives at $\mathcal{X}_{0}$. In the ideal case $\alpha = \beta$, which implies that moving agents reach the attractive point with a null speed. Fig.\ref{fig:fig3} shows the proposed parametric function $\vec{G}_{near}(r)$ for approximating velocity fields in a near range of interactions over a two-dimensional plane $(x,y)$. For visualization purposes, it is considered $\mathcal{X}_{0} = \left[ \begin{array}{c} x_{0} \\ y_{0} \end{array} \right]$, $x_{0} = 0$, $y_{0} = 0$, $\sigma^{2} = 15$, $\beta = 1$ and $\alpha = 1$.

Each velocity field in a near range of interactions can be described by the set of parameters $\xi = [\beta,\alpha, \mathcal{X}_{0}, \sigma]$. For estimating each of them, it is proposed an iterative gradient descent method explained in detail on section \ref{LearningEnv}.

As can be seen in equation \eqref{eq3}, since $\beta$ and $\alpha$ are always positive, it is assumed that moving agents decrease their speeds while they approach the attractive areas. Accordingly, the proposed model applies to moving agents that tend to remain on the attractive center of forces, i.e., their speeds tend to be null near it. By taking into consideration the latter assumption, it is possible to define a distance $r_{switch}$, from which agents start decreasing their velocity at approaching to $\mathcal{X}_{0}$.

In this sense, the model described in equation \eqref{eq3} is valid for $r \leq r_{switch}$. When distances between agents' locations and attractive areas are greater than $r_{switch}$, another model must be proposed to characterize the agents' dynamics in terms of attractive forces. Under this perspective, the following section introduces a way to model agents' velocities when they are located at far distances from $\mathcal{X}_{0}$.  

\subsection{Far range of interaction}\label{FarModels}

When moving agents are located at a distance $r>r_{switch}$ from an attractive area, it is proposed to use a probability density function (PDF) to describe the agents' velocities until they start reducing their speeds due to a near range of interactions effect previously explained in section \ref{NearModels}. From this viewpoint, the model of a far range of interaction works until agents present a continuous deceleration through time, which indicates they start approximating to an attractive area.  

PDFs are proposed to characterize interactions at far distances from attractive centers of force due to their properties for representing stochastic phenomena. It is hypothesized that agents' velocities at far ranges of interaction can be slightly affected by multiple factors different from the effect of attractive forces in the environment. Nonetheless, at this stage it is assumed a predominant agents' velocity due to their interactions with an attractive area, i.e., it is assumed a main constant influence on the agents' velocity due to a particular destination point (goal) that makes them move towards a center of force $\mathcal{X}_{0}$.  

In this work, it is proposed to represent the velocity's magnitude of an agent localized in $r>r_{switch}$ as the $L^2$ norm among the velocities experienced in each spatial axis, such that $||V_{\mathcal{X}_{far}}||_2$, where $V_{\mathcal{X}_{far}}$ represents the agent's velocity in a single location measured in a far range of interaction. Based on that, the set of all velocities' magnitudes due to an attractive area in a far range of interaction can be written as: 
$$||V_{far(set)}||_2 = \{||V_{\mathcal{X}_{far,1}}||_2, ||V_{\mathcal{X}_{far,2}}||_2,...,||V_{\mathcal{X}_{far,J}}||_2\},$$
where $J$ is the total number of measurements taken in a far range of interaction. Since a main constant velocity is assumed during this stage, it is defined a random variable $\gamma$ that distributes according to the log-normal distribution $ln\mathcal{N}(\mu,\sigma_{far}^2)$ that is adjusted based on $||V_{far(set)}||_2$. Such distribution allows to represent variables that are only defined for positive values and that can be concentrated mainly in one value, as it is the case of velocities' magnitudes in far range of interactions. From this viewpoint, velocity fields in such stage can be written as shown in equation \eqref{eq4}: 
\begin{equation}\label{eq4}
\vec{G}_{far}(r) = \gamma \, \hat{r}.
\end{equation} 
as explained before, $\gamma$ is a random variable that distributes in accordance a log-normal distribution. After characterizing near and far interactions between agents and an attractive area of the environment, it is possible to define a full effect $\vec{G}^m(r)$ produced by a particular area $m$, such as shown in equation \eqref{eq5}:
\begin{equation}\label{eq5}
\vec{G}^m(r) =
\begin{cases} 
	\vec{G}_{near}^m(r) & r \leq r_{switch} \\
	\vec{G}_{far}^m(r) & r > r_{switch}. 
\end{cases}
\end{equation} 

Consequently, as different agents move through an environment, a total set of fields $\vec{G}^{M}(r)$ can be obtained, where $M$ is the number of identified attractive areas that have been characterized inside a scene. From this perspective, each effect can be seen as a letter that is part of a vocabulary which is learned hierarchically as new agents move inside an environment. The next section describes a methodology that aims at the incremental learning of environment properties based on moving agents' observations. 

\section{Hierarchical environment representation}\label{SemanticRep}
The present section explains a methodology by which effects generated by static attractive areas can be modeled as control parameters into an initial Kalman filter (KF) formulation from which effects of interactive areas can be extracted and then modeled into it. It is also shown how trajectories that are globally nonlinear in terms of space can be incrementally grouped into segments in which agents' dynamics preserve a constant orientation. Each identified segment is here explained by an attractive velocity field whose effect is modeled through the switching formulation described in \ref{ProblemDef}. Finally, it is proposed that each trajectory can be seen as a sequence of attractive areas acting on the moving agents. This process can be considered as the activation of different letters in the environment (each one associated to an attractive area) that gives a semantic representation of activities in the environment.  

\subsection{Kalman filter modeling}	\label{KFmodel}
Supposing location measurements of single agents in an environment are defined as $Z_k$ for each instant of time $k$ and assuming that such measurements are equally time spaced, i.e., $\Delta k$ remains constant, it is possible to define the agent state vector $X_{k}$ as its position and velocity at $k$. Accordingly, let measurements and state of agents be related by the expression $Z_{k} = HX_{k} + \nu_{k}$, where $\nu_{k}$ represents a zero-mean normal distributed noise due to errors on the measurements taken at time $k$ and $H$ is a matrix that maps the actual state space $X_{k}$ onto the observations $Z_{k}$.

Since the proposed method aims at the hierarchical learning of agents' temporally non-linear motions, initially, a random walk model centered on agents' observations is hypothesized as the most general way to represent unknown dynamical behaviors. Based on this, observations of moving agents that deviate from a random walk model can be used to incrementally explain effects of attractive areas in a particular location $Z_{k}$ from which the deviation was observed. 

Dynamics of agents are modeled by a bank of Kalman filters that grows sequentially as new patterns associated with attractive areas in the environment are revealed by looking at agents' motions through time. Accordingly, equation \eqref{eq6} shows the initial KF dynamical model based on a random walk model from which the bank of filters is built up afterwards.
\begin{equation}\label{eq6}
X_{k + 1} = FX_{k} + w_{k},
\end{equation} 
where
$$ X_k=\left[ \begin{array}{c} \mathcal{X}_{k} \\ \dot{\mathcal{X}_{k}} \end{array} \right], \quad F=\begin{bmatrix} I_{n} \\ 0_{n,n}  \end{bmatrix},$$
$w_{k}$ is a zero-mean normal distribution associated with the random noise of the proposed dynamic model and $n$ represents the number of observed dimensions of the environment as shown in equation \eqref{eq1a}.
 
Let $X_{k|k -1}^0$ be the prediction of a KF based on the random walk model described in equation \eqref{eq6} given the updated state estimation at the time $k-1$, i.e., $X^0_{k-1|k-1}$. It is assumed that significant deviations from the KF's predictions carry information about effects of unknown attractive areas. Accordingly, when the measurement $Z_k$ arrives, it is possible to compute $\tilde{Y}^0_{k}$, called innovation of measurement residual, which is defined as shown in equation \eqref{eq7}: 
\begin{equation} \label{eq7}
\tilde{Y}_{k }^0 = Z_{k} - H X_{k|k-1}^0,
\end{equation}
where $X_{k|k-1}^0 = FX_{k-1|k-1}^0$ and it is defined as the random walk KF's prediction. In general, innovations can be seen as a quantity that measures the deviation that a proposed dynamical model presents respecting observations. In the ideal case, the value $\tilde{Y}_{k}^0$ would tend to zero, which means that the proposed model, based on a random walk dynamics, can explain the observations of agents in the environment correctly. Following this reasoning, in cases where innovations are significantly different from zero, the dynamical model should be modified in order to describe more accurately the observed agents' dynamics. To do so, it is considered a modification in random walk model, the new model formulation is shown in equation \eqref{eq8}:
\begin{equation} \label{eq8}
X_{k+1} = FX_{k} + BU_{k} + w_{k},
\end{equation}  
where a control parameter $BU_{k}$ is introduced in order to obtain innovations that tend to zero. $U_{k}$ can be seen as a velocity contribution that makes an agent follow certain motion when it is located at the coordinates $Z_{k}$. $B$ is a matrix that maps such velocity contribution into the agent's state, such that $B =\left[ \begin{array}{c} \Delta k I_{n} \\ I_{n} \end{array} \right]$.

Similarly to the random walk model, innovations of the proposed new model are defined as $\tilde{Y}_{k}^1 = Z_{k} - H X_{k|k-1}^1$. Accordingly, in order to make $\tilde{Y}_{k}^1 = 0$, it is proposed to use the innovations produced by the random walk model, $\tilde{Y}_{k}^0$, into the control parameter $BU{k}$ of the proposed new model. By taking into consideration that $X_{k|k-1}^1 = F X_{k-1|k-1}^1 + BU_{k-1}$, it is possible to rewrite the expression of innovations generated by the new model as shown in equation \eqref{eq9}:
\begin{equation} \label{eq9}
Z_{k+\Delta k} - H (F X_{k-1|k-1}^1 + BU_{k-1}) = 0.
\end{equation}  

By substituting the updated state produced by the KF based on the proposed new model with the updated state found with the random walk based model, such that $X_{k-1|k-1}^1 := X_{k-1|k-1}^0$, it is possible to rewrite equation \eqref{eq9} as shown next:
\begin{equation} \label{eq9a}
\underbrace{(Z_{k+\Delta k} - H F X_{k-1|k-1}^0)}_{\substack{\text{Random walk KF's} \\ \text{innovation:} \tilde{Y}_{k}^0}} - HBU_{k-1} = 0
\end{equation} 
from which it is possible to infer that: 
\begin{equation} \label{eq10}
\tilde{Y}_{k}^0 = H B U_{k-1}  \Longrightarrow U_k=\frac{\tilde{Y}^0_{k}}{\Delta k}\bigg\} \text{Valid in } Z_k.
\end{equation}

As can be seen in equation \eqref{eq10}, the control vector $U_{k}$ is a velocity quantity related to the effect that attractive forces exert over moving agents in the particular measurement $Z_k$. For this viewpoint, since different agents localized in the same coordinated point $Z_k$ can be motivated by different attractive areas, each velocity field effect should be individuated and included into the KFs' dynamic models that compose the proposed bank of filters as shown \eqref{eq11}:    
\begin{equation} \label{eq11}
A_s : = (X_{k+\Delta k} = FX_{k} + BU_{k}^s + w_{k}),
\end{equation} 
where $U_{k}^s = \frac{\tilde{Y}^0_{k,s}}{\Delta k}$, $s$ is the indexation of each potential attractive area and $A_s$ corresponds to the dynamical model associated to each of them. In this sense, each $A_s$ represents a new KF model built from the innovations produced by the random walk model described in \eqref{eq6}. At the moment of defining an attractive model $A_s$, information about innovations generated by the previously characterized set of dynamical models $\{A_1, A_2...,A_{s-1}\}$ should be taken into consideration in order to merge similar velocity field models. In this work, a merging  process of similar velocity fields is considered as a post processing step and a classic k-means algorithm is used to group similar fields. From this viewpoint, the proposed bank of KFs is created incrementally by observing deviations from a reference random walk model. Each learned new model includes information about attractive properties of the environment.

In order to relate the proposed attractive field described in equation \eqref{eq5} to the control inputs $U_k^s$, it is necessary to identify parts on a trajectory in which velocity effects produced by radial attractive force fields are valid. For this purpose, it is considered a method for segmenting trajectories explained in section \ref{TrajSeg}.

\subsection{Trajectory segmentation} \label{TrajSeg}

Since it is hypothesized that movements of agents are produced by static areas that exert radial attractive force fields, it is expected that trajectories performed by agents are spatially quasilinear and pointing every time at the direction of a main attractive area. Nevertheless, in real cases, motions of agents do not have the same orientation in every moment. However, they can be divided into quasilinear segments on which the assumption of interactions with attractive radial fields is valid. 

\begin{figure}[h!]
		\includegraphics[width=8cm]{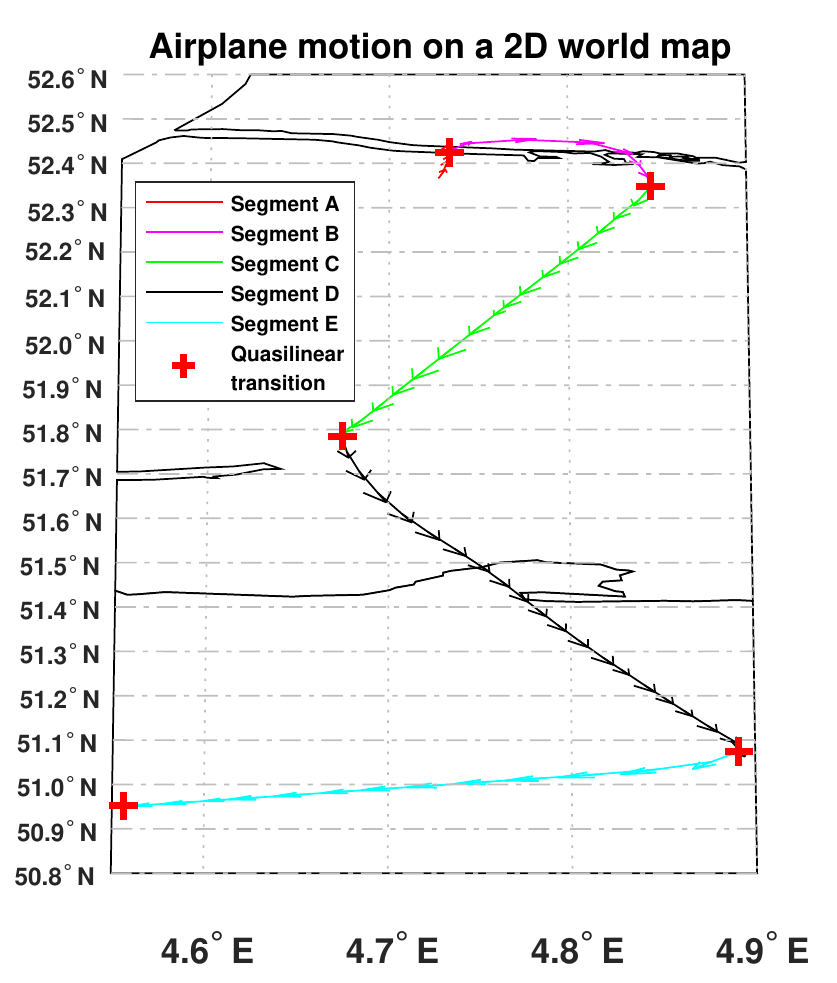}
	\caption{\csentence{Segmentation of spatially nonlinear trajectories.}
		Partition of a trajectory into quasilinear segment is shown in a trajectory performed by an aircraft.}
	\label{fig:fig4}
	\end{figure}

The possibility of dividing a trajectory into quasilinear segments can be associated with the presence of multiple motivations that an agent follows as time evolves. In other words, a moving agent can be attracted by multiple areas that can be seen as dynamical motivations that change in time as the agent moves through the environment reaching multiple goals. An example of this behavior can be clearly identified in airplane trajectories, where aircrafts perform different approaches to intermediate points during a flight. Fig.\ref{fig:fig4} depicts an example of such behavior for a commercial flight that goes from Amsterdam to Brussels on a Latitude-Longitude plane. 

In Fig.\ref{fig:fig4}, spatial quasilinear segments in the trajectory are identified with different colors. Red crosses show the points in which transitions between different quasilinear segments take place. Accordingly, from Fig.\ref{fig:fig4}, it is possible to see how multiple quasilinear paths can explain a single trajectory in terms of spatial transitions. Accordingly, each segment identified in Fig.\ref{fig:fig4} is proposed to be explained by an attractive area that exerts a radial force field over the environment. 

Angles produced by the estimated velocities $U_k$ are used in order to segment the trajectory into quasilinear motions in an online way. Since the control vector $U_{k}$ can be seen as the agent's velocity due to the interaction with external forces when at observing $Z_{k}$, the orientation of this vector can be used as a measurement of alignment with attractive centers of force. If an agent is approaching to an attractive area in the environment, it is expected that orientations of $U_{k}$ do not change significantly through time. Based on this, it is proposed to define a window time $t_w$ in which directions of velocities $U_k$ remain stable. Let $W_a$ be the windowed version of the velocities $U_k$ such that:
$$W_k = \{U_{k}, U_{k+1},U_{k+2},..., U_{k+a}\},$$
where $a$ is the number of samples that fit in an interval of time $t_w$. Once a number of $a$ velocities $U_{k}$ is obtained, the window data $W_k$ is used to approximate the parameters of a von Mises distribution $\mu_{\theta}$ and $\kappa_{\theta}$. This distribution is selected due to its properties of representing the direction of entities in a probabilistic way, $\mu_{\theta}$ represents distribution's mean and $\frac{1}{\kappa_{\theta}}$ is analogous to the variance in a normal distribution.   

Defining $\hat{\mu}_{\theta,k}$ and $\hat{\kappa}_{\theta,k}$ as the estimated parameters of the von Mises distribution based on the window data $W_k$, it is proposed a deviation angle $\theta_{dev}$ from which it is possible to distinguish when an agent moving aligned to a particular attractive area. Accordingly, when an accumulated probability higher than $0.9$ is obtained in the range $[\hat{\mu}_{\theta,k}-\theta_{dev}, \hat{\mu}_{\theta,k}+\theta_{dev}]$, it is hypothesized that the agent is attracted to an area. Velocity contributions calculated in following times, i.e., from $U_{k+a+1}$ onwards are compared with the current von Mises distribution through a Mahalanobis distance $D_M$ \cite{Mahalanobis1936}, in order to evaluate whether it belongs to the previously estimated distribution of if it is necessary to hypothesize a new one. For doing so, a distance limit $D_{\theta}$ is introduced, such that if a number of $n_{\theta}$ velocities contributions produce a distance $D_M > D_{\theta}$, the process is reinitialized with a new window $W_{k+a^{*}+1}$, where $a^{*}$ is the minimum window size $a$ together with the following velocity contributions that were classified as part of the von Mises distribution estimated from the window $W_k$.

In this work, each quasilinear segment, indexed with the letter $s$, is associated with the presence of an attractive area in the environment, see equation \eqref{eq11}. Accordingly, once a segment is detected, a velocity field approximation is iteratively performed based on an attractive force hypothesis. Algorithm \ref{algo1} describes the proposed sequential process for identifying quasilinear motions in agents' dynamics and approximating velocity fields from them. 
\begin{algorithm}[h!]
	
	\label{algo1}
	
	\KwData{\\
		* [$Z, U$] Measurements with their respective velocity contributions.\\
		* [a] Minimum window size.\\ 
		* [$\theta_{dev}$] Tolerance angle for considering quasilinear agents' dynamics.\\
		* [$D_{\theta}$] Maximum distance to distiguish a new velocity contribution from a probability distribution.\\	
		* [$n_{\theta}$] Number of velocity contributions that do not belong to a probability distribution before proposed a new segmentation.\\	
		* [$s$] Segment (attractor) counter \\
		* [$traj_{stable}$] Indicator of alignment in the agent's motion\\
		* [$n_{\Theta}$] Counter of velocity contributions that do not belong to the current probability density function \\
		* $[I_{s}, F_{s}]$ Initial and final measurements' indexes corresponding to the segment (attractor) $s$.
	}
	\KwResult{\\
		* Velocity field estimations based on attractivive influences. \\ \textbf{Procedure:} }
	$N\gets$ size of $Z$\\	
	\If{$N == a$}{
		$\theta_{U}\gets$ orientations of $U$\\
		$[\hat{\mu}_{\theta}, \hat{\kappa}_{\theta}]\gets$ von Mises parameters based on $\theta_{U}$ \\
		\If{$CDF([\hat{\mu}_{\theta} - \theta_{dev},\hat{\mu}_{\theta} + \theta_{dev}])>0.9$}{
			$s =  s + 1$\\
			$I_{s} \gets$ assignation of first index of [$Z, U$] \\			
			$traj_{stable}\gets TRUE$  \\
			$n_{\Theta} = 0$\\
			Velocity field estimation related to attractor $s$ 			
		}
		\Else{
			Redefined [$Z, U$] with their last $(a-1)$ values\\		
		}
	}
	\Else{
		\If{$stableTrajectory == TRUE$}{
			$D_M \gets$ Mahalanobis distance between last $\theta_{U}$ and current von Mises probability distribution \\
			\If{$D_M < D_{\theta}$}{
				$[\hat{\mu}_{\theta}, \hat{\kappa}_{\theta}]\gets$ von Mises parameters based on $\theta_{U}$ \\			
				Velocity field estimation related to attractor $s$ 	
			}\Else{
			$traj_{stable}\gets FALSE$\\
			$n_{\Theta} = n_{\Theta} +1$ \\
			\If{$n_{\Theta} == n_{\theta}$}{
			$F_{s} \gets$ assignation of last index of [$Z, U$] \\	
			Remove all the elements [$Z, U$] \\
			}
		}
	}
}
Wait new observation and append it to $Z$ and $U$ \\
Execute Segmentation/Field approximation Algorithm
\caption{Segmentation/Field approximation.}
\end{algorithm}

Some algorithms for segmenting signals into linear components and reconstruction of trajectories were identified. However, for the particular problem addressed in this work, where moving agents are influenced by a main goal, information about the direction and magnitude of their velocities should be taken into consideration to make a correct segmentation in terms of attractive forces. The number of segments identified with this approach corresponds to the quantity of potential attractive areas in the environment. A list of the principal characteristics of identified algorithms for piecewise linear segmentation and reconstruction of trajectories are compared with the current proposal in table \ref{table:table1}.         

\begin{table*}[ht]	
	\caption{Comparison among strategies for segmenting and reconstructing signals.} \label{table:table1}
	\begin{tabular}{ccccc}
		\hline
		Method used in  	& \begin{tabular}{@{}c@{}}Segmentation/Reconstruction \\ Type \end{tabular} 								& Input features  		& Error measure	& Output\\ \hline
		\cite{Ge2001}		& Batch (Top-Down) 		& \begin{tabular}{@{}c@{}} Data from \\ etch machine\end{tabular}  												& \begin{tabular}{@{}c@{}}$\ell^2$ distance \\ between line and data points \end{tabular}								& \begin{tabular}{@{}c@{}} Identification of \\states in a HMM \end{tabular} \\
		\cite{Keogh2004}	& \begin{tabular}{@{}c@{}}Online \\ (Sliding window + Top down) \end{tabular}  								& Data points series 	& \begin{tabular}{@{}c@{}}$\ell^2$ distance \\ between line and data points \end{tabular}   							& \begin{tabular}{@{}c@{}} Signal's piecewise \\ linear representation \end{tabular}\\ 
		\cite{Zhou2012}		& \begin{tabular}{@{}c@{}}Online \\ (Slope change threshold) \end{tabular}  								& Data points series	& \begin{tabular}{@{}c@{}}Distance between \\ reference slopes \end{tabular} 																		& \begin{tabular}{@{}c@{}} Signal's piecewise \\ linear representation \end{tabular}\\		
		\cite{Hunter1999}	& Batch (Bottom-Up) 	& \begin{tabular}{@{}c@{}} High frequency \\ data points \end{tabular}  														& \begin{tabular}{@{}c@{}}$\ell^2$ distance between \\  line and super-interval data \end{tabular}						& Identification of events\\
		\cite{Liu2008}		& \begin{tabular}{@{}c@{}} Online (Furthest candidate \\ segmenting point) \end{tabular} 								& Data points series 	& \begin{tabular}{@{}c@{}} Maximum vertical \\ distance to a line \end{tabular}										& \begin{tabular}{@{}c@{}} Signal's piecewise \\ linear representation \end{tabular}\\
		\cite{Liu2008} 		& \begin{tabular}{@{}c@{}} Online \\ (Backward segmentation) \end{tabular} 								& Data points series    & \begin{tabular}{@{}c@{}} Maximum vertical \\ distance to a line \end{tabular}										& \begin{tabular}{@{}c@{}} Signal's piecewise \\ linear representation \end{tabular}\\
		\cite{Teixeira2011} & \begin{tabular}{@{}c@{}} Offline (nonlinear \\ state estimation) \end{tabular}  								& \begin{tabular}{@{}c@{}} Aircraft \\ sensor data \end{tabular}  														& Root-mean-square error																				& \begin{tabular}{@{}c@{}} Aircraft path \\reconstruction \end{tabular} \\
		\cite{Garcia2015} 	& Offline  				& \begin{tabular}{@{}c@{}} Interactive Multiple \\ Mode filter data \end{tabular}   													& \begin{tabular}{@{}c@{}} Distance to \\ predefined  modes of flight \end{tabular}	& \begin{tabular}{@{}c@{}} Identification of modes\\  of flight \end{tabular} \\
		\begin{tabular}{@{}c@{}}Proposed \\work \end{tabular} 
							& \begin{tabular}{@{}c@{}}Online (Sliding window \\  and angle distributions) \end{tabular}  				
													& \begin{tabular}{@{}c@{}} Velocity vectors \\ from bank of filters\end{tabular}  	& \begin{tabular}{@{}c@{}}Mahalanobis distance \\ between angle and distribution \end{tabular}
																							&  \begin{tabular}{@{}c@{}} Velocity vectors linked\\ to an attractive field\end{tabular} \\ \hline
	\end{tabular}
	
\end{table*}

From table \ref{table:table1}, it can be seen that the proposed method presents the advantage of being online. additionally, it uses the dynamics of agents as inputs for performing the segmentation process. Since produced segments are assigned to central force fields, the agents' velocity angles result to be an important feature for detecting alignment with goals that agents desire to reach by taking the shortest possible path.  

It is possible to see from algorithm \ref{algo1} that the final result of the proposed method consists of a set of velocity fields related to attractive areas identified by quasilinear segments in the agents' motions. Accordingly, as a segment grows in time, an iterative estimation of the velocity field's characteristics is performed based on the model described in sections \ref{NearModels} and \ref{FarModels}. Section \ref{LearningEnv} explains in detail how this process is performed as new agents' observations are obtained.

\subsection{Learning of environment properties}\label{LearningEnv}

Let the pair of vectors $[Z_k,U_k]$ be the agent's observations with their respective control vectors obtained by the KF based on the random walk model proposed in equation \eqref{eq6}. Additionally, suppose that $[Z_k,U_k]$ are acquired incrementally such that algorithm \ref{algo1} is applied as new observations arrive such that $[Z_k^s,U_k^s]$ are vectors that belong to a particular segment $s$, i.e., they are produced due to an unique attractor.  

Since parameters $U_k^s$ are related to velocity components that deviate agents from a random walk behavior, it is proposed to use them in order to fit the velocity field model presented in equation \eqref{eq5}. Accordingly, for distinguishing between far and near ranges of interaction, magnitudes of $U_k^s$ are analyzed through consecutive time instants, such that: If $||U_k^s||_{2}$ values decrease continuously in a range of time $\Delta K_{switch}$, the moving agent is assumed to be in a near range of interaction, proposed in equation \eqref{eq3}. For this case,  its measurements and control vectors will be labeled as $[Z_{k(near)}^s,U_{k(near)}^s]$. Otherwise, the moving agent is assumed to follow the far range of interaction model described in equation \eqref{eq4}, in such case, its measurements and control vectors will be labeled as $[Z_{k(far)}^s,U_{k(far)}^s]$.

In order to characterize the velocity field in a far range of interaction associated to a segment $s$, the magnitudes $||U_{k(far)}^s||_{2}$ are used to estimate the random variable $\lambda_s$, such as proposed in equation \eqref{eq4}.

Estimating the center of force $\mathcal{X}_{0}$ in a far range of interaction is not possible since there is not conclusive information about velocity decrements. Nonetheless, it is feasible to determine a line on which the center of force belongs. Such line is defined from the last observation $Z_k$ and has a slope determined by the mean of $\lambda_s$. 

$[Z_{k(near)^s}, U_{k(near)}^s]$ are used to estimate the unknown parameters in equation \eqref{eq3} for a particular attractive area $s$. The measurement and control input from which next $||U_k^2||_2$ values start decreasing are labeled as $[Z_{k(switch)}^s, U_{k(switch)}^s]$. The magnitude of $U_{k(switch)}^s$ is taken as an estimation of to the term $\beta$ of equation \eqref{eq3}, such that $\hat{\beta}_s= ||U_{k(switch)}^s||_2$. Additionally, it is assumed $\hat{\alpha}_s = \hat{\beta}_s$, such that moving agents tend to remain on the attractive center of force when they arrive in it. Then, an iterative gradient descent is proposed to minimize the error between magnitudes $||U_{k(near)}^s||_2$ and the proposed velocity field formulation of equation \eqref{eq3} based on the sparse space representation of the observed data $Z_{k(near)}$.

Accordingly, the function to optimize is the mean square error between $||U_{k(near)}^s||_2$ and the proposed velocity field function described in \eqref{eq3} with the fixed values of $\alpha = \hat{\alpha}_s$ and $\beta = \hat{\beta}_s$. By assuming a total number of $N$ measurements with their control inputs identified in a near attractive range of interaction for an area $s$, i.e., $[Z_{k(near)^s}, U_{k(near)}^s]$ with $k$ from 1 to $N$, the function to optimize is shown in equation \eqref{eq12}:
\begin{equation} \label{eq12}J(\sigma, \mathcal{X}_{0}) = \frac{1}{N}\sum_{k=1}^{N} (||U_{k(near)}^s||_2 - G(r_{near}))^2.
\end{equation} 

Since $G(r_{near})$ is a differentiable function respecting the parameters $\sigma$ and $\mathcal{X}_{0}$, a classical gradient descent algorithm is performed by taking the partial derivative of the function $J(\sigma,\mathcal{X}_{0})$ respecting each of them. Both expressions are shown in equation \eqref{eq13}:
\begin{equation} \label{eq13}
\begin{split}
\qquad \qquad \frac{\partial J}{\partial \mathcal{X}_{0}} &= \frac{4\alpha}{N\sigma^2} \sum_{k=1}^{N} \psi_k e_k r_{k},
\\
\quad \frac{\partial J}{\partial \sigma} &= \frac{4\alpha}{N\sigma^3}  \sum_{k=1}^{N} \psi_k e_k  r_{k}^2,\quad\quad\quad
\end{split}
\end{equation} 
where
\begin{gather*} 
\qquad r_{k} = (Z_{k(near)} - \mathcal{X}_{0}),\quad e_k =\alpha e^{- \frac{r^2_k}{\sigma^2}},\\
\qquad \qquad \qquad \psi_k =  U_{k(near)} + e_k - \beta.
\end{gather*}

As a moving agent is approaching an attractive center of force, the number $N$ of measurements identified in a near range of interaction will increase as more vectors $[Z_{k(near)}^s,U_{k(near)}^s]$ are available. In other words, each time that $N$ increases, new estimations $\hat{\sigma}_s$ and $\hat{\mathcal{X}}_{0,s}$ made by gradient descent process are obtained. Supposing that a total number of $Q$ estimations of $\hat{\mathcal{X}_0}$ and $\hat{\sigma}$ have been obtained by the multiple computations of the gradient descent method, it is proposed to fuse all of them for a given segment $s$, i.e., $[\hat{\mathcal{X}}_{0,s,q},\hat{\sigma}_{s,q}]$ where $q$ goes from $1$ to $Q$, by performing a weighted average over the $Q$ estimations such as shown in equation \eqref{eq14}: 
\begin{equation} \label{eq14}
[\hat{\mathcal{X}}_{0,s(avg)},\hat{\sigma}_{s(avg)}] = \sum_{q=1}^{Q}\big([\mathcal{X}_{0,s,q},\sigma_{s,q}] w_{1,q} w_{2,s,q} \big), 
\end{equation} 
where $w_{1,q}$ is a weight related to the number of samples with which the gradient descent was executed in the iteration $q$. Each $w_{1,q}$ is normalized respecting the maximum number of samples with which the gradient descent have been performed. $w_{2,s,q}$ is a weight related to the error $J_{s,q}(\beta, \alpha, \sigma, \mathcal{X}_0)$ obtained on each iteration $q$ at estimating the properties of the attractor associated to the segment $s$. $w_{2,s,q}$ values are normalized respecting the lowest error obtained.    

The pair of parameters $[\hat{\mathcal{X}}_{0,s(avg)},\hat{\sigma}_{s(avg)}]$ obtained by applying equation \eqref{eq14}, is the final estimation of the velocity field parameters [$\mathcal{X}_0$,$\sigma$] related to the effects of an attractive area based on the segment $s$. $\hat{\mathcal{X}}_{0,s(avg)}$ represents the estimated location of the attractive center of force associated to based the segment $s$ and $\hat{\sigma}_{s(avg)}$ encodes information about the shape of its effect in the environment.

By replacing the vector of parameters' estimations $\hat{\xi}_s = [\hat{\beta}_s ,\hat{\alpha}_s, \hat{\mathcal{X}}_{0,s(avg)}, \hat{\sigma}_{s(avg)}]$ in equation \eqref{eq3}, it is possible to obtain an expression that explains the effect of an attractive area in a near range interactions, such that, $\hat{\vec G}_{near}^s(r) = \vec G_{near}(\hat{\xi}_s)$. Similarly, for a far range of interaction, it is possible to write $\hat{\vec G}_{far}^s(r) = \hat{\lambda}_s$. Consequently, a final estimation of attractive effects can be defined as:
$$
\hat{\vec{G}}^s(r_s) =
\begin{cases} 
\hat{\vec{G}}^s_{near}(r_s) & r\leq r_{s(switch)} \\
\hat{\vec{G}}^s_{far}(r_s) & r> r_{s(switch)}, 
\end{cases}
$$ 
where $r_{s(switch)}$ represents the distance from which agents start decreasing their velocities at approaching the attractor and $r_s = ||Z_k - \mathcal{X}_{0,s}||_2$ is the distance between an agent and its center of force. Estimated velocity fields $\hat{\vec{G}}^s(r_s)$ can be used to rewrite equation \eqref{eq11} in terms of the proposed parametric function such as shown next: 
\begin{equation} \label{eq15}
A_s : = X_{k+\Delta k} = FX_{k} + B \hat{\vec{G}}^s(r_s) + w_{k}.
\end{equation}  
 
As explained before, the purpose of this paper is to estimate the velocity fields $\vec{G}^s(r_s)$ generated by attractive areas in environments. From equation \eqref{eq15}, it is possible to see that a bank of filters composed of the effects of attractive fields related to each segment $s$ is obtained by applying the proposed methodology. From this viewpoint, each KF dynamical model $A_s$ can be seen as a letter that is learned on the fly as patterns in agents' motions are detected. Consequently, as multiple letters are observed and characterized, it is possible to build a vocabulary in terms of learned environment properties. In the next section, some results at modeling such properties in synthetic and real trajectories are shown and discussed. 

\section{Results}\label{Results}
The proposed method was tested in two different scenarios: i) Synthetic data, in which an environment with different attractive areas is proposed and whose characteristics are inferred based on agents that move towards different goals. ii) Pedestrian dataset, where dynamics of people moving through an indoor place are analyzed and attractive points in the environment are identified.

\subsection{Synthetic data}\label{SimulatedData}
\begin{figure}[h!]
		\includegraphics[width=8cm]{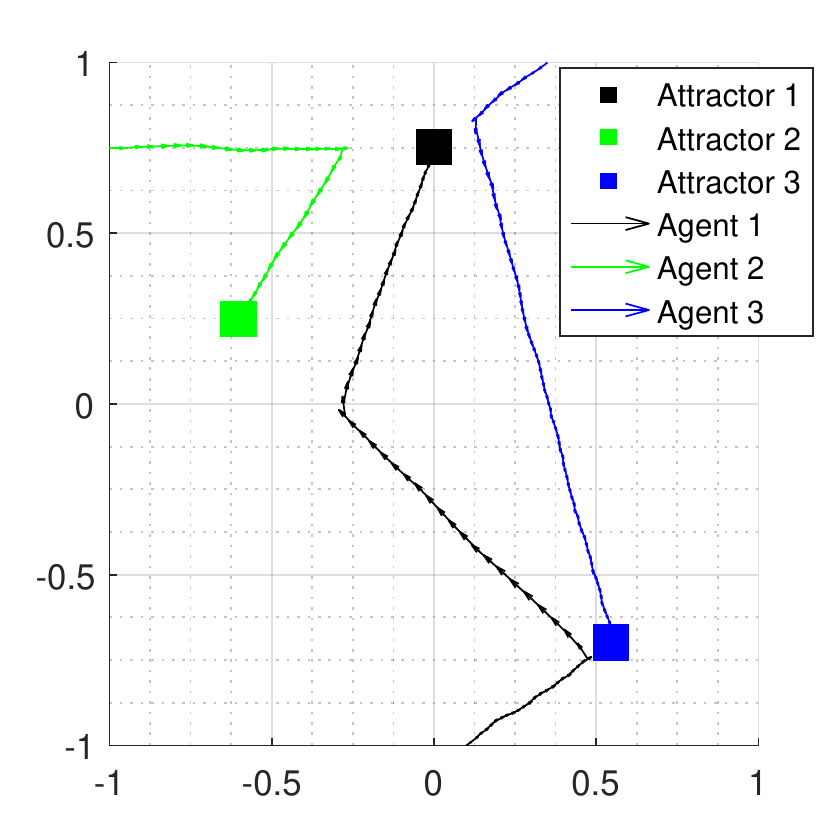}
	\caption{\csentence{Synthetic data layout.}
		A scenario composed three attractive zones is proposed to test the method.}
	\label{fig:fig5}
    \end{figure}

In order to validate the proposed method, a simulated scenario in which moving agents are motivated by different attractive zones is considered. Three radial attractors that influence moving agents are proposed as shown in Fig.\ref{fig:fig5}. 

It is proposed that agents interact with at least one attractor before they reach their final destination. In this sense, situations in which agents change dynamically their destinations (goals) are simulated and analyzed in an online way. This simulation assumes that agents always start moving from a random side of the environment such as shown in Fig.\ref{fig:fig5}. Noise from a uniform distribution $\mathcal{U}(-b,b)$ is considered in order to affect the motion of agents in directions $x$ and $y$. The value of $b$ is selected according to the speed that agents present each instant of time, such that it is adjusted to a specific signal-to-noise ratio $\textrm{SNR} = ||\vec{V}_{\mathcal{X}}||_2/b$, where $\vec{V}_{\mathcal{X}(t)}$ is the velocity exhibited  by an agent in a location $\mathcal{X}$. Attractive areas' effects are simulated according to the formulation shown in equation \eqref{eq3} with $\beta  = -\alpha$. The selected values of the parameters for each attractor are shown in the first column of table \ref{table:table2}.

It was considered a total of 150 trajectories that start randomly from one side of the environment and whose final/intermediate(s) points of attraction are also selected at random. The proposed method was applied to the trajectories in order to recognize attractive effects in the environment. 

\begin{figure}[h!]
	\includegraphics[width=8cm]{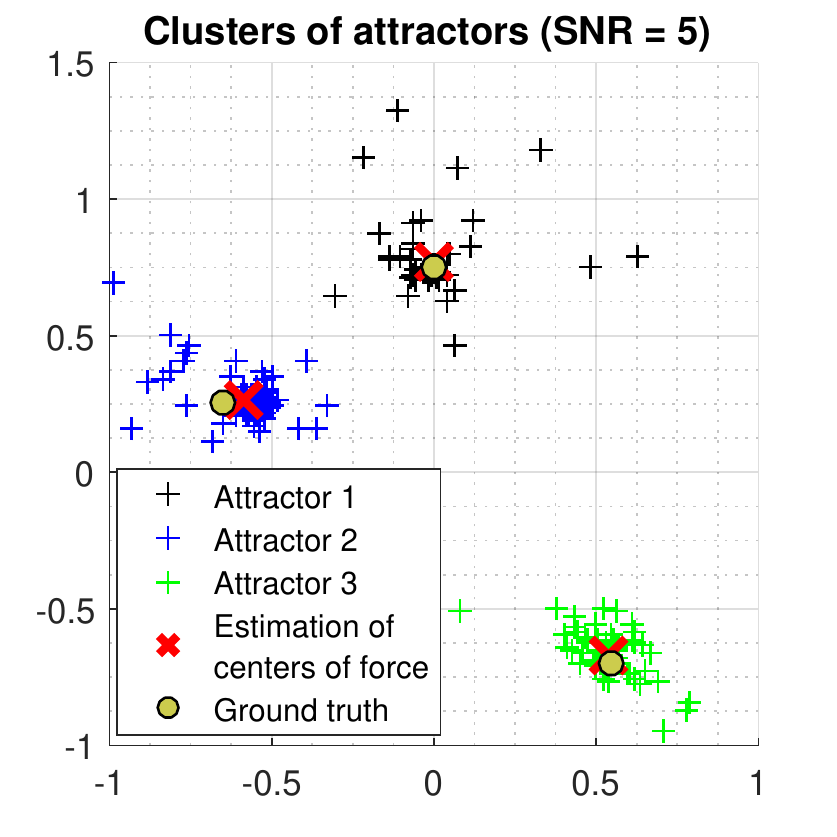}
	\caption{\csentence{Clusters of simulated data.}
		Result of attractive clusters based on synthetic trajectories.}
	\label{fig:fig6}
\end{figure}

By taking the estimations of attractive centers of force generated by agents that behave according to a near range of interaction explained in \ref{NearModels}. It is possible to obtain a map with different locations of attractive points such as shown in Fig.\ref{fig:fig6}. As can been seen, similar centers of force can be clustered by using a classic k-means algorithm in order to identify attractive areas in the scene that combine information of similar attractors' estimations.

From Fig.\ref{fig:fig6}, it can be seen that each identified cluster corresponds to an attractive zone. Crosses are calculated by taking the average position of all the velocity fields that belong to a particular cluster. Similarly, the average of $\sigma$ and $\beta$ components of velocity fields is taken in order to characterize the properties of each merged attractor. 

Attractors resulting from the clustering process can be seen as letters that conform a vocabulary that characterizes the environment. Accordingly, the parameters associated with each letter can be expressed as shown in equation \eqref{eq17}: 
\begin{equation} \label{eq17}
\xi_{(merg),m} = \frac{1}{P_m}\sum_{p_{m} = 1}^{P_{m}}(\hat{\xi}_{p_{m}}), 
\end{equation} 
where $m$ indexes each identified cluster composed by similar estimated fields that in turn are indexed as $p_m$. $\hat{\xi}_{p_{m}}$ represents the estimated parameters $\hat{\xi}_s = [\hat{\beta}_s ,\hat{\alpha}_s, \hat{\mathcal{X}}_{0,s(avg)}, \hat{\sigma}_{s(avg)}]$ coming from a segment $s$ and relabeled according to the clusters (letters) reached with the k-means algorithm.

In order to evaluate the algorithm performance, it is possible compare the parameters estimated for each merged attractor with the respective ground truth values. Accordingly, table \ref{table:table2} shows the values estimated at different SNRs for each attractor. 

\begin{table}[h!]
	\caption{Estimated attractive parameters for different SNRs.} \label{table:table2}
	\centering
	\begin{tabular}{cccc}
		\hline
		\begin{tabular}{@{}c@{}}Attractor\\ ground truth \end{tabular}& $\textrm{SNR} = 10$  & $\textrm{SNR} = 6$  & $ \textrm{SNR} = 1.5$\\ \hline
		$\begin{bmatrix} x_1 = 0  	\\ y_1 = 0.75  	\\ \beta_1 = 0.09 	\\ \sigma^2_1 =  0.1 \end{bmatrix} $ & $\begin{bmatrix}  -0.0022  	\\     0.7716  	\\           0.0976 \\           0.1558 \end{bmatrix} $ & $\begin{bmatrix}  -0.0025  	\\     0.7823  	\\           0.1019	\\           0.1897 \end{bmatrix} $ & $\begin{bmatrix}   0.0028   \\     0.7385  	\\           0.0361 \\           0.0163 \end{bmatrix} $\\
		$\begin{bmatrix} x_2 = -0.6 \\ y_2 = 0.25	\\ \beta_2 = 0.108 	\\ \sigma^2_2 =  0.2 \end{bmatrix} $& $\begin{bmatrix}  -0.599  	\\     0.2528  	\\           0.109 	\\           0.1923 \end{bmatrix} $& $\begin{bmatrix}  -0.6029  	\\     0.2536  	\\           0.1154 \\           0.2207 \end{bmatrix} $& $\begin{bmatrix}  -0.5748  	\\     0.2634  	\\           0.0067 	\\       0.0029 \end{bmatrix} $\\
		$\begin{bmatrix} x_3 = 0.55	\\ y_3 = -0.7  	\\ \beta_2 = 0.117 	\\ \sigma^2_3 =  0.3 \end{bmatrix} $& $\begin{bmatrix}  0.5587  	\\    -0.7121  	\\           0.133 	\\           0.3706 \end{bmatrix} $& $\begin{bmatrix}  0.5596  	\\    -0.7147  	\\           0.138 	\\           0.3887 \end{bmatrix} $& $\begin{bmatrix}  0.539  	\\     -0.675  	\\           0.0042	\\           0.0011 \end{bmatrix} $\\ \hline
	\end{tabular}
\end{table}

From table \ref{table:table2}, it can be seen that estimations of centers of force locations are not affected drastically at the presence of noise. However, it is observed a significant change in the estimations $\beta$ and $\sigma$ when SNR is close to 1. Such behavior indicates that at low SNRs the main direction of trajectories is preserved but the magnitudes of their velocities are significantly affected.     

In order to illustrate how the overall performance in the estimation of $\hat{\xi}_s = [\hat{\beta}_s ,\hat{\alpha}_s, \hat{\mathcal{X}}_{0,s(avg)}, \hat{\sigma}_{s(avg)}]$ at different SNRs, it is proposed to take the following expression shown in equation \eqref{eq18}:
\begin{equation} \label{eq18}
\epsilon_{\tau} = \frac{\varphi_\tau}{\max (\varphi_1,\varphi_2,...,\varphi_T)},
\end{equation} 
where
$$
\varphi_{\tau} = sum\{ \vartheta_\tau \}, \quad \vartheta_\tau = \frac{\lambda_\tau}{\gamma},
$$
$$\gamma = \max(\{\lambda_1, \lambda_2,..., \lambda_T \}),\quad \lambda_\tau = sum\{\varrho_\tau\},$$
$$\varrho_\tau = \{\rho_{1,\tau}, \rho_{2,\tau}, ..., \rho_{M,\tau} \}, \quad \rho_{m,\tau} = |\xi_m -  \hat{\xi}_{m,\tau}|,$$
$m$ and $\tau$ index the attractive zones and the different SNRs respectively. The total number of attractors and SNRs are represented with $M$ and $T$. $\xi_m$ represents the ground truth of the parameters that define the attractor $m$. $\hat{\xi}_{m,k}$ is the estimation of attractive parameters related to an attractor $m$ at a specific SNR $k$. 

\begin{figure}[h!]
		\includegraphics[width=8cm]{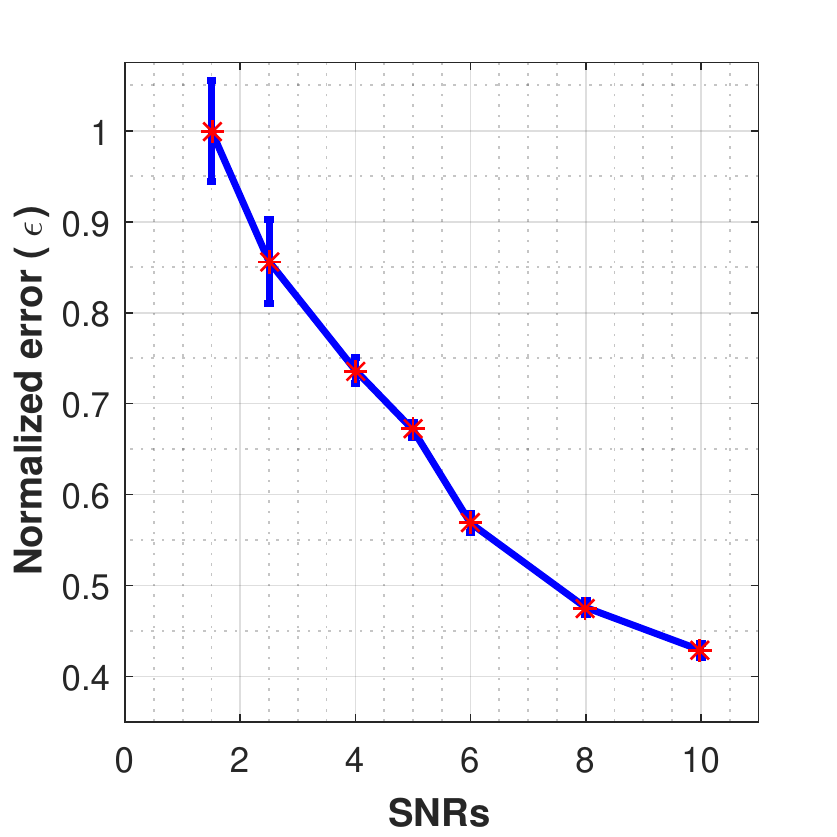}
	\caption{\csentence{Performance of simulated data.}
		The normalized error produced with the proposed method is shown for different SNR values.}
	\label{fig:fig7}
\end{figure}

By taking the normalized error expressed in equation \eqref{eq18}, it is possible to obtain the performance results shown in Fig.\ref{fig:fig7} for different signal-to-noise ratios. Variances produced by estimations of the three attractive zones are plotted as error bars for each SNR case.  

From results obtained with simulated data, it is possible to see that the proposed approach recognizes and characterizes radial attractive fields with different intensities and shapes at several SNR values by looking at the motion of individuals. Additionally, the proposed method does not need a large number of data vectors and parameterizes areas according to a known function. This represents a an advantage in comparison with our previous approach \cite{Damian2016} that needs trajectories all around the environment for estimating correctly the effects of force fields in environments. For validating our method on real data, it is proposed to analyze a dataset composed by pedestrians' trajectories in an indoor environment.    

\subsection{Pedestrian dataset}\label{PedestrianDataset}
\begin{figure}[h!]
		\includegraphics[width=8cm]{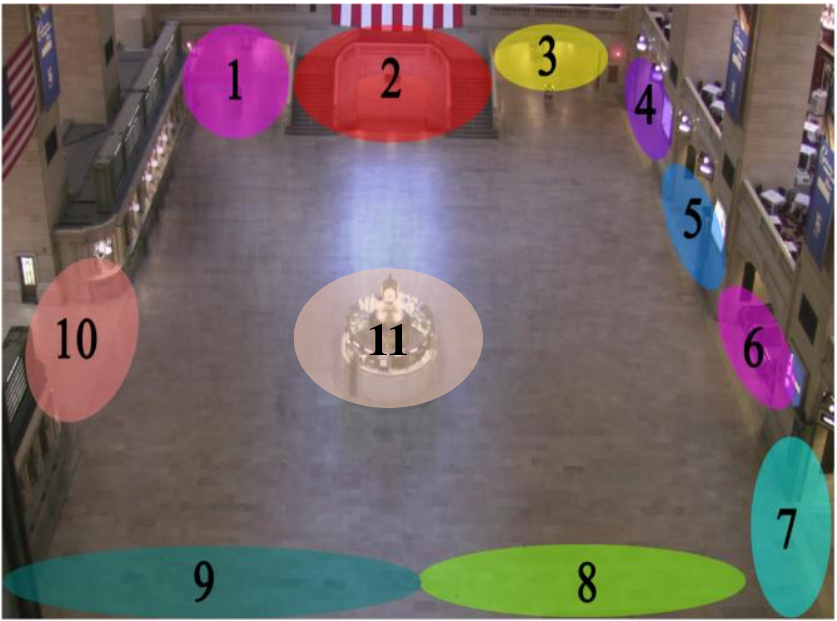}
	\caption{\csentence{Zones of interest in pedestrian database.}
		10 source/destination zones identified by \cite{Yi2015} plus a center zone of interaction.}
	\label{fig:fig8}
\end{figure}

The dataset proposed by Yi in \cite{Yi2015} was used to test our approach. The database is composed by 12,684 pedestrian trajectories in an indoor place where each individual is manually labeled from a one-hour crowd surveillance video. In their work, they identify 10 zones that act as source/destination points for pedestrians and it is manually labeled in Fig.\ref{fig:fig8}, as can be seen, an 11th zone is introduced as the center point of the scene that is hypothesized to be an attractor for some pedestrians that interact with it. Consequently, the plot shown in Fig.\ref{fig:fig8} can be seen as a map of attractors in the scene. By using our approach, it is expected to identify those points in terms of their location and the effect they produce on moving pedestrians. 

\begin{figure}[h!] 
	\includegraphics[width=8cm]{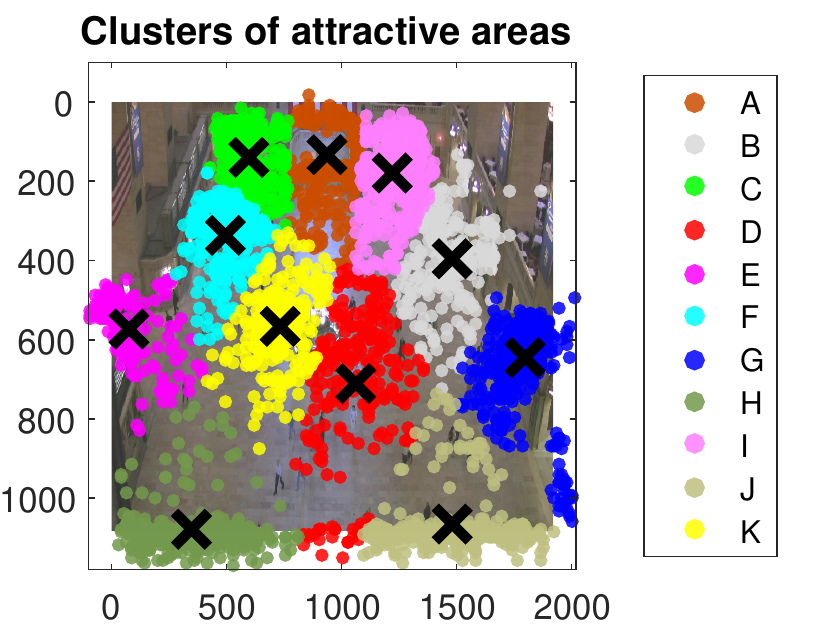}
	\caption{\csentence{Clustering of attractive center of forces.}
		similar estimated center of points are grouped to characterize attractive areas.}
	\label{fig:fig9}
\end{figure}

By grouping similar estimations of attractive centers of force generated by pedestrians that behave according to a near range of interaction, it is possible to obtain a map of the attractive zones' locations such as shown in Fig.\ref{fig:fig9}. Similar to the case of synthetic data, parameters that belong to the same attractive zone are averaged based on equation \eqref{eq17} to obtain the parametrization of each merged version of attractive fields in the scenario. Estimations of attractive zones' locations are depicted as red crosses on Fig.\ref{fig:fig9}.

\begin{figure}[h!] 
		\includegraphics[width=8cm]{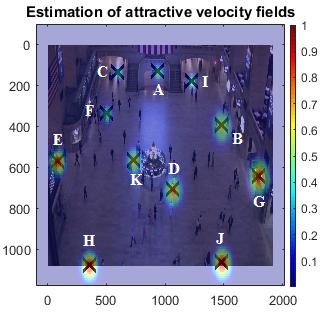}
	\caption{\csentence{Characterized attractive areas in the environment.}
		A representation of attractive areas in terms of their intensity, location and shape is depicted on the environment.}
	\label{fig:fig10}
\end{figure}

In order to visualize the presence and effect of each attractive zone, it is proposed a map that is shown in Fig.\ref{fig:fig10}. Areas with null values (blue colors) represent the absence of attractive fields. Zones with larger values (red colors) represent the presence of attractors that where agents feel more attracted in terms of intensity. The shape of fields are visualized as the radius around each center of force (black cross) until the color map becomes totally blue. Consistently, the larger such radius is, the near range of interaction effect becomes broader, i.e., a larger period of deceleration is detected.    

Results obtained in this database demonstrate that from the dynamics of moving agents it is possible to infer the locations and characteristics of attractive areas. Particularly, qualitative results can be seen by comparing our approximation of attractive centers of force shown in Fig.\ref{fig:fig10} with the zones manually labeled by \cite{Yi2015} and depicted in Fig.\ref{fig:fig8}. It can be seen that zones labeled as $\{C, A, I, B, G, J, H, E\}$ in Fig.\ref{fig:fig10} correspond respectively to the labels $\{1, 2, 3, 5, 6, 7, 8, 9, 10\}$ in Fig.\ref{fig:fig8}.

Since there is not a ground truth available to verify the intensity or shape of attractors associated to $\beta$ and $\sigma$ values respectively, the only comparison that can be done it is terms of centers of force localizations between Fig.\ref{fig:fig8} and Fig.\ref{fig:fig10} as proposed before. 

	\begin{figure}[h!] 
		\includegraphics[width=8cm]{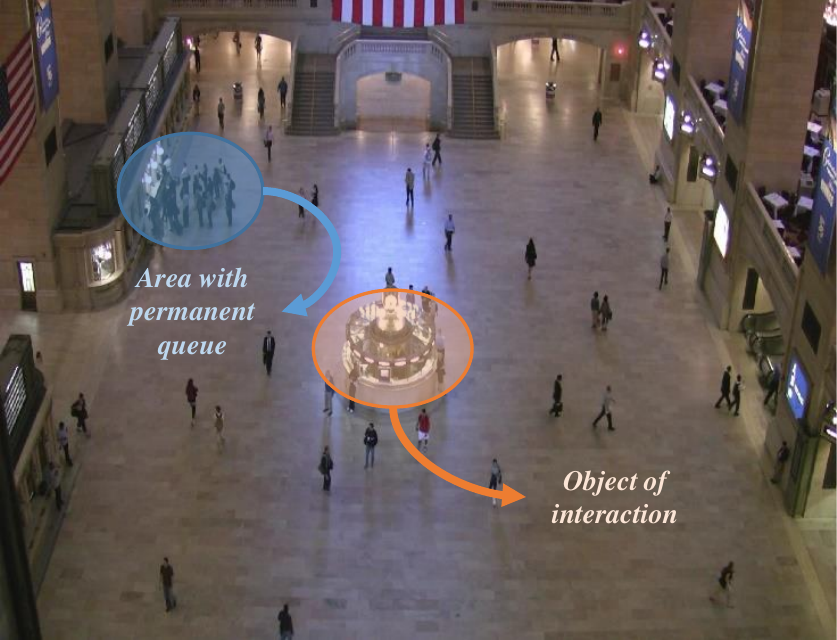}
	\caption{\csentence{Characteristics of the environment.}
		Areas that become relevant according to the context of the environment.}
	\label{fig:fig11}
\end{figure}
Additionally, it was identified the area labeled with letter $F$ corresponds to an effect of a queue that pedestrians follow through the video. Zones identified as $K$ and $D$ are related to pedestrians' interactions with the central object of the scene. Both, queue and object's effect are illustrated in Fig.\ref{fig:fig11}. According to that, it is demonstrated that the proposed methodology is capable of recognizing effects derived from contextual situations in the environment by looking at movements of dynamical agents through time.    

\section{Conclusions}\label{Conclusions}
A method for characterizing attractive forces in statical environments based on movements of agents is proposed. To do that, it is put forward a methodology based on the minimization of innovations from a Kalman filter based on a random walk model. Innovations from this model are used in order to the explain external effects that affect agents' motions. Information that encodes properties of the environment effects on agents is proposed to be added as a control input into the initial random walk formulation.  

An incremental learning of environment properties is considered. Characteristics of attractive zones around observed agents are learned on the fly. A method for individuating attractive fields effect based on radial force hypothesis is proposed based on a quasilinear segmentation of the agents' dynamics. An iterative method that approximates velocity fields produced by attractive zones is proposed when only information about agents' dynamics is available.  

Obtained maps can be used to perform further analysis by following a semantic approach where each characterized attractive area (letter) can be considered as part of a vocabulary, if activation sequences of attractive areas are detected with the proposed bank of filters, it would be possible to extract information about situations in an environment. 

For future work, it is proposed to adopt a semantic representation of learned static areas properties, such that more complex interactions between agents and environments can be characterized by applying the built bank of filters based on attractive effects to new trajectories. For this approach, it is planned to using the innovations with their covariance matrices produced by the built bank of filters to characterize repulsive effects in environments. 

\section{Competing interests}\label{Interests} 
The authors declare that they have no competing interests of any type. 

\section{Funding}\label{Funding} 
This work was partially developed within the MIE (Mobilità Intelligente Ecosostenibile) project, co-funded by the Italian Ministry of University and Research for the National Technological Cluster for the Smart Communities.

\section{Authors' contributions}\label{Contributions} 
All the authors have participated in writing the manuscript and have revised the final version. All authors read and approved the final manuscript. 


\begin{backmatter}
	
%
%
%
	
	\bibliographystyle{bmc-mathphys} 
	\bibliography{bmc_article}      

\end{backmatter}
\end{document}